# Balance Scene Learning Mechanism for Offshore and Inshore Ship Detection in SAR Images

Tianwen Zhang, Xiaoling Zhang, Jun Shi, Shunjun Wei, Jianguo Wang, Jianwei Li, Hao Su, and Yue Zhou

*Abstract*—Huge imbalance of different scenes' sample numbers seriously reduces Synthetic Aperture Radar (SAR) ship detection accuracy. Thus, to solve this problem, this letter proposes a Balance Scene Learning Mechanism (BSLM) for offshore and inshore ship detection in SAR images. BSLM involves three steps: 1) based on unsupervised representation learning, a Generative Adversarial Network (GAN) is used to extract the scene features of SAR images; 2) using these features, a scene binary cluster (offshore/inshore) is conducted by K-means; 3) finally, the small cluster's samples (inshore) are augmented via replication, rotation transformation or noise addition to balance another big cluster (offshore), so as to eliminate scene learning bias and obtain balanced learning representation ability that can enhance learning benefits and improve detection accuracy. This letter applies BSLM to four widely-used and open-sourced deep learning detectors, i.e., Faster Regions-Convolutional Neural Network (Faster R-CNN), Cascade R-CNN, Single Shot multi-box Detector (SSD) and RetinaNet, to verify its effectiveness. Experimental result on the open SAR Ship Detection Dataset (SSDD) reveal that BSTM can greatly improve detection accuracy, especially for more complex inshore scenes.

## I. INTRODUCTION

SHIP detection in Synthetic Aperture Radar (SAR) images is attracting increasing scholars' attention [1]-[21] for its great value in traffic control, salvage at sea, fishery management, etc.

So far, many traditional SAR ship detection methods have been proposed [1]-[4]. Hou *et al.* [1] designed a multilayer Constant False Alarm Rate (CFAR) SAR ship detector. Zhu *et al.* [2] proposed a template-based SAR ship detection method. Xie *et al.* [3] and Zhai *et al.* [4] also proposed saliency-based methods. However, the above traditional feature extraction methods have huge difficulty in detecting more complex inshore ships.

In recent years, increasing scholars [9]-[21] have started to make extensive research into deep learning-based SAR ship detection methods, e.g., 1) based on Faster R-CNN [5], Cui *et al.* [9] proposed a dense attention pyramid network to detect multi-scale ships, and Lin *et al.* [10] also proposed a squeeze and excitation Faster R-CNN to improve detection accuracy; 2) based

This work was supported in part by the National Natural Science Foundation of China under Grant 61571099, 61501098 and 61671113, and in part by the National Key R&D Program of China under Grant 2017YFB0502700. *(Corresponding author: Xiaoling Zhang.)*

T. Zhang, X. Zhang, J. Shi, S. Wei, J. Wang and H. Su are with the School of Information and Communication Engineering, University of Electronic Science and Technology of China, Chengdu 611731, China (e-mail: twzhang@std.uestc.edu.cn; xlzhang@uestc.edu.cn; shijun@uestc.edu.cn; weishunjun@uestc.edu.cn; wang_jg@uestc.edu.cn; suhao@std.uestc.edu.cn).

J. Li is with the 3rd Graduate Student Team, Naval Aeronautical University, Yantai 264000, China (e-mail: lgm_jw@163.com).

Y. Zhou is with the School of Electronic Information and Electrical Engineering, Shanghai Jiao Tong University, Shanghai 200240, China (e-mail: sjtu_zy@sjtu.edu.cn).

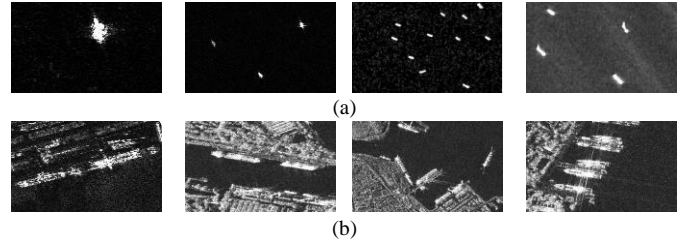

(a)

(b)

Fig. 1. Ships in SAR images. (a) Offshore scenes; (b) Inshore scenes.

on Cascade R-CNN [6], Wei *et al.* [11] designed a high-resolution SAR ship detection network, and Li *et al.* [12] used a cascade smoothing operator to ease speckle noise in SAR images; 3) based on SSD [7], Wang *et al.* [13] applied transfer learning to improve accuracy, and Yang *et al.* [14] fused multi-scale feature for multi-scale ship detection; 4) based on RetinaNet [8], Wang *et al.* [15] performed ship detection in Gaofen-3 images, and Liu *et al.* [16] proposed an improved loss function to enhance training benefits. However, these deep learning methods still pay less attention to complex inshore scenes, and thus have a huge accuracy gap between inshore scenes and offshore ones.

One possible reason is the huge imbalance of offshore scene and inshore one in sample number. Such huge imbalance refers that offshore samples in Fig. 1(a) are far more than inshore ones in Fig. 1 (b) among most datasets. As a result, it brings a huge imbalance in learning representation ability, causing lower inshore scene accuracy than offshore one. Moreover, our previous work [17] found that the detection accuracy of inshore scenes is too poor as well, seriously hindering our further progress.

Thus, to solve this problem, this letter proposes a novel Balance Scene Learning Mechanism (BSLM) for offshore and inshore SAR ship detection. BSLM involves three steps: 1) using unsupervised representation learning, a GAN [22] is used to automatically extract different scenes' features in an unsupervised manner by the confrontation between a generator and a discriminator; 2) using extracted features, a scene binary cluster (inshore/offshore) is conducted by K-means; 3) finally, the sample number of the small cluster (inshore samples) is augmented by replication, rotation transformation or noise addition to balance another big cluster (offshore samples), so as to eliminate different scenes' learning bias and achieve balanced learning representation ability of different scenes, which can enhance learning benefits. To confirm BSLM's effectiveness, we apply it to four widely-used and open-sourced detectors, i.e., two high-accurate but low-speed two-stage detectors (Faster R-CNN and Cascade R-CNN) and two high-speed but low-accurate one-stage detectors (SSD and RetinaNet). Experimental results on the open SSDD dataset [12] reveal that BSTM can greatly improve detection accuracy, especially for more complex inshore scenes. Notably, BSLM is a universal mechanism that is also useful for other object detectors and other SAR datasets, of great value.

The main contributions of this letter are as follows:
1) BSLM is proposed for offshore and inshore SAR ship detection via the balance idea of different scenes' sample number.
2) GAN is used to extract the scene features of SAR images to differentiate between offshore ships and inshore ships based on the unsupervised representation learning.

## II. METHODOLOGY

### A. Motivation of BSLM

#### 1) Motivation 1: Online Hard Example Mining

Shrivastava *et al.* [23] proposed Online Hard Example Mining (OHEM) to excavate hard samples where the samples with big training losses are seen as hard ones which are accumulated into a pooling. Once the sample number in the pooling equals a training batch, they will be regarded as a new training batch to be trained by networks repeatedly. In general, inshore samples can be seen as hard ones and the offshore as easy ones. However, OHEM is sensitive to noise labels [24] that widely exist in most SAR datasets, so it is not rather suitable for SAR ship detection, so it is essential to excavate difficult inshore samples for emphatical training to enhance their learning representation ability.

#### 2) Motivation 2: Balance Learning

Pang *et al.* [24] found three imbalances in object detection, i.e., positive-negative sample, feature and object imbalances. We also found SAR image scene imbalance, i.e., offshore samples often account for a larger proportion than inshore ones in whole dataset, which is in line with almost SAR ship datasets, and also seems to accord with the fact that ocean area of the earth is much larger than land, but such scene imbalance can bring a learning bias, i.e., detectors can obtain strong learning ability in offshore scenes for more offshore samples, but poor one in inshore scenes for less inshore samples, because more data can bring better learning effects generally. Finally, leaving aside the differences in scene complexity, just considering the big gap in scene sample number, the accuracy of the inshore inevitably is inferior to the offshore, given its poor learning representation ability [See Fig. 2(a).], so it is essential to balance scene sample number in order to obtain balanced scene learning representation. Finally, BSLM is proposed to make the above two imbalances in Fig. 2(a) to be balanced again [See Fig. 2(b).].

### B. Implementation of BSLM

#### 1) Step 1: Unsupervised Scene Feature Extraction

In Fig. 2, the core of BSLM is to distinguish different scenes. In fact, it is a challenging task because dataset's publishers did not provide real scene class labels (inshore/offshore), so this is an unsupervised process. It is not advisable to adopt traditional methods to extract scene features for limited generalization performance and excessive human intervention. Fortunately, GAN [22], a modern promising unsupervised algorithm [25], makes this problem solved smoothly. So far, by such unsupervised representation learning, Lin *et al.* [25] and Radford *et al.* [26] have applied GAN to scene feature extraction of optical images, so we also adopt GAN to extract the SAR images scene features. Moreover, it is also feasible to collect some samples as offshore ships with homogeneous features to differentiate inshore ships in a supervised manner, but it is time-consuming and laborious, and may also cause limited application in other more datasets.

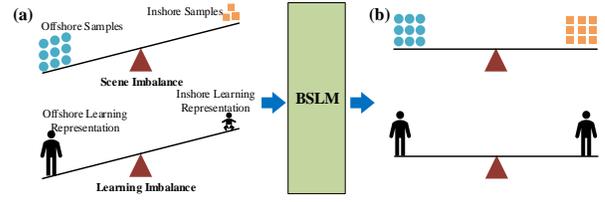

Fig. 2. Scene imbalance and learning imbalance.

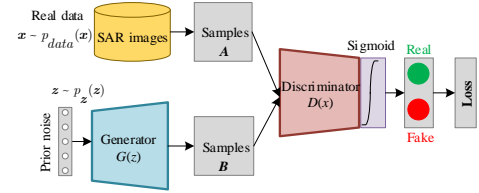

Fig. 3. Model of GAN.

#### a). Model of GAN

Fig. 3 shows the model of GAN. From Fig.3, a GAN consists of a generator $G(z)$ and a discriminator $D(x)$. The input of the generator $G(z)$ is a random prior noise $z$ with a distribution $p_z(z)$ that is used to: 1) learn the generator's distribution $p_g$ on the real data $x$; 2) represent a mapping to data space as $G(z; \theta_g)$ where $G$ is a differentiable function of a network with parameters $\theta_g$. The output of the generator $G(z)$ is the generated samples $B$. The inputs of the discriminator $D(x)$ are both the samples from the real data $x$ with a distribution $p_{data}(x)$ and the generated samples $B$ of the generator. The output of the discriminator $D(x)$ is a single scalar $D(x; \theta_d)$ that is used to represent the probability that $x$ came from real samples $A$ rather than fake samples $B$, by using a sigmod function defined by $y = 1/(1+e^{-x})$.

In training, the generator $G(z)$ generates fake samples $B$ that come from its learning on the real samples $A$ meanwhile the discriminator $D(x)$ strives to distinguish fake samples $B$ from real samples $A$. Then, based on the feedback of the training loss function, the generator $G(z)$ will learn harder to make fake samples $B$ closer and closer to real samples $A$, so as to deceive the discriminator $D(x)$, but the discriminator $D(x)$ will also work harder to improve its identification ability so as to resist the cheating of the generator $G(z)$. Finally, the above confrontation process between the generator $G(z)$ and the discriminator $D(x)$ enables GANs to accurately learn the real data distribution [22]. Moreover, GAN can also learn multi-resolution scene features, for the existence of multi-resolution images in the training set.

#### b). Network Architecture of GAN

Similar to Radford *et al.* [26], we establish the GAN, shown in Fig. 4. From Fig. 4, there are 7 deconvolutional blocks in the generator (C1~C7) and 7 convolutional blocks in the discriminator (C8~C14). The network input of the generator is set as a 100-dimension random noise vector obeying a uniform distribution. After a series of deconvolution operations, it is mapped into an output image with 256×256×3 dimension. Then, the real samples $A$ and generated samples $B$ are used as inputs of the discriminator network. After a series of convolution operations, they are mapped into a feature vector with 1024×4×4 dimension.

Similar to Lin *et al.* [25], we concatenate the last 3 convolution blocks' feature maps (C12, C13 and C14) in the discriminator to form the scene feature vector of SAR images $F$, in order to aggregate the mid- and high-level information [25], i.e.,

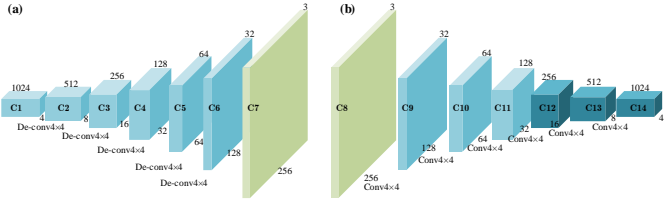

Fig. 4. Network architecture of GAN. (a) Generator. (b) Discriminator.

$$F = \text{flatten}\left(F_{C12} \copyright \text{MaxPool}^{2\times}(F_{C13}) \copyright \text{MaxPool}^{4\times}(F_{C12})\right) \quad (1)$$

where $F_{C12}$, $F_{C13}$ and $F_{C14}$ respectively denotes the feature maps of C12, C13 and C14, $\copyright$ denotes concatenate operation, Max-Pool$^{2\times}$ and MaxPool$^{4\times}$ respectively denotes 2 times and 4 times max pooling, and flatten is used to reshape feature maps into a column vector, i.e., from $\mathbb{R}^{1792\times4\times4}$ to $\mathbb{R}^{28672\times1\times1}$ where 1792 equals 256+512+1024 and 28672 equals 1792×4×4. Finally, the scene feature vector of SAR images,

$$F = (f_1, f_2, f_3, \cdots, f_{28671}, f_{28672})^T \quad (2)$$

is clustered using K-means algorithm.

#### c). Training of GAN

We train $D(\boldsymbol{x})$ to maximize $D(G(\boldsymbol{z}))$ that means assigning correct labels to real samples and samples from $G(\boldsymbol{z})$, meanwhile train $G(\boldsymbol{z})$ to minimize $\log[1-D(G(\boldsymbol{z}))]$. Moreover, the weights of $G(\boldsymbol{z})$ are fixed when training $D(\boldsymbol{x})$, and the weights of $D(\boldsymbol{x})$ are fixed when training $G(\boldsymbol{z})$. Finally, $D$ and $G$ play the two-player minimax game with value function $V(G, D)$ [22]:

$$\min_G \max_D V(D, G) =$$
$$\mathbb{E}_{\boldsymbol{x}\sim p_{data}(\boldsymbol{x})}[\log D(\boldsymbol{x})] + \mathbb{E}_{\boldsymbol{z}\sim p_z(\boldsymbol{z})}[\log(1-D(G(\boldsymbol{z})))] \quad (3)$$

More details about GAN can be found in [22], [25], [26].

#### 2) Step 2: Scene Binary Cluster

Using the scene feature vector $F$, K-means is used to conduct a scene binary cluster. Fig. 5 shows the clustering results of K-means on the training set of SSDD. In Fig. 5, according to our observation, almost all inshore samples can be correctly classed into the small cluster with less samples, meanwhile almost all offshore samples can be correctly classed into another big cluster. One possible reason is that GAN could have chosen better features to represent different scenes based on an even a small number of samples of the offshore and inshore classes.

We conduct a performance evaluation of clustering using the cluster internal criteria [30] (i.e., real scene class labels are not required.), as is shown in Table I. In Table I, Calinski-Harabaz-Index is used to measure the compactness within clusters defined by Calinski *et al.* [27] (A higher value means better performance.), Davies-Bouldin-Index is used to measure the average value of maximum similarity of clusters defined by Davies *et al.* [28] (A lower value means better performances and its value range is [0,1]) and Silhouette-Coefficient is used to measure the density within clusters and the dispersion between clusters defined by Rousseeuw *et al.* [29] (A higher value means better performance and its value range is [-1,1]). In Table I, the high Calinski-Harabaz-Index, small Davies-Bouldin-Index and high Silhouette-Coefficient jointly reveal the good cluster performance using the scene features from GAN by K-means, from the perspective of statistical significance of datasets.

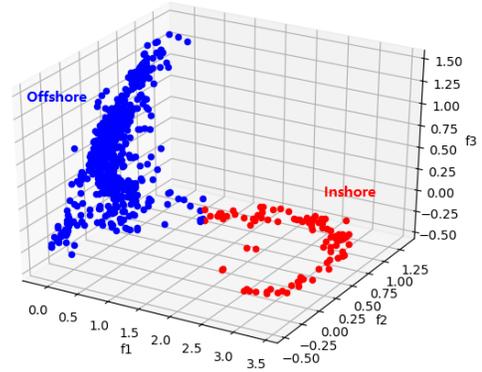

Fig. 5. Clustering results of K-means on the training set. Visualization presentation of $(f_1, f_2, f_3)$ from Eq. (2) for more intuitive observation.

TABLE I
EVALUATION INDICES OF K-MEANS CLUSTER RESULTS

| Type | Value |
|---|---|
| Calinski-Harabaz-Index | 2991.61 |
| Davies-Bouldin-Index | 0.26 |
| Silhouette-Coefficient | 0.80 |

Moreover, we also conduct a performance evaluation of clustering based on the cluster external criteria [30] (i.e., real scene class labels are required.). However, SSDD's publisher [12] did not provide the real labels of inshore or offshore, so we specifically make scene class labels according to human visual experience, where the samples containing lands are regarded as the inshore scene and the samples not containing lands are regarded as the offshore one. Given above, the average accuracy of scene cluster reaches 96.15%, showing the scene features extracted from GAN are rather effective.

Finally, we regard the small cluster as inshore scene and another big one as offshore scene. In fact, our such practice is also certainly in line with the actual application situation, because, in almost all existing SAR ship datasets, the number of inshore samples is universally less than that of offshore ones, which also seems to accord with the fact that ocean area of the earth is also much larger than land.

#### 3) Step 3: Inshore Sample Augmentation

Finally, we augment the small cluster with less samples (inshore samples) via sample replication, rotation transformation or noise addition to make the sample number of two clusters basically equal so as to obtain a balance between the inshore and the offshore. Finally, the learning bias of different scenes is eliminated and the learning representation of different scenes is balanced, bringing better learning benefits.

### III. EXPERIMENTS

#### A. Dataset

We use the first open SSDD dataset [12] to verify our work. There are 1,160 SAR images with 500×500 average size from Sentinel-1, TerraSAR-X and RadarSat-2 in SSDD[1], with various polarizations, various resolutions and abundant ship scenes.

---

[1] In our previous work [17], SSDD is divided into a training set, a validation set and a test set by 7:2:1, but such random partition may cause accuracy randomness on the test set, so Li *et al.* [12] appeal to use 8:2 ratio as a training set and a test set (image indexes' suffix 1 and 9 as the test set). Furthermore, we also revised the error labels of the test set and the revised test set is available on: https://pan.baidu.com/s/1YvsrP84l_At-svoZu44q-w (password: e5fn).

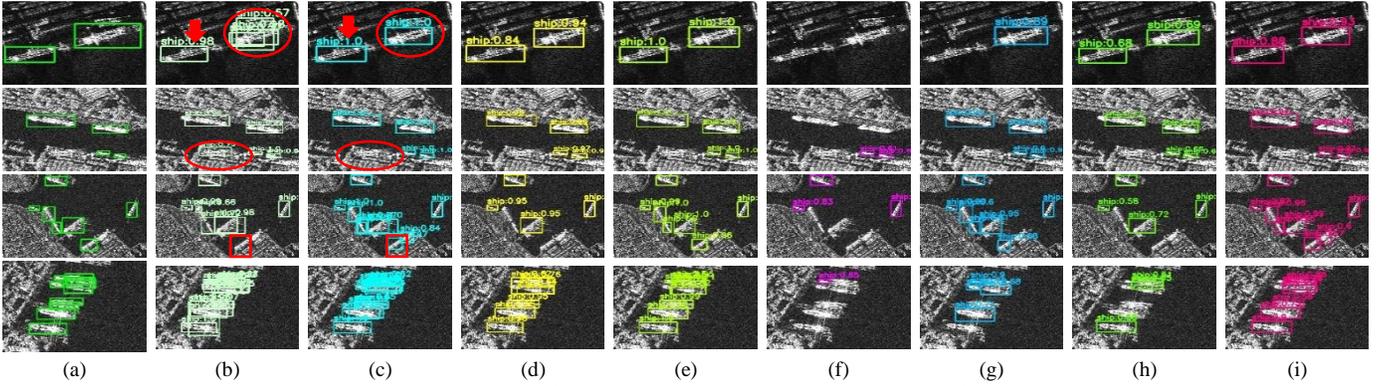

Fig. 6. Detection results on the test set. (a)Real ships; (b)Faster R-CNN; (c)Faster R-CNN+BSLM; (d)Cascade R-CNN; (e)Cascade R-CNN+BSLM; (f)SSD;(g) SSD+BSLM; (h)RetinaNet; (i)RetinaNet+BSLM. False alarm rates $P_f$ of (b)-(i) are 16.49%,10.05%,5.84%,7.05%,3.57%,6.18%,5.74%,15.72% ($P_f$ = 1-Precision).

TABLE II
EVALUATION INDICES OF DIFFERENT METHODS WITH BSLM OR WITHOUT BSLM.

| Type | Method | BSLM | Inshore + Offshore | | | Inshore | | | Offshore | | |
|------|--------|------|------|------|------|------|------|------|------|------|------|
| | | | Recall | Precision | mAP | Recall | Precision | mAP | Recall | Precision | mAP |
| Two-Stage | Faster R-CNN | ✗ | 89.34% | 83.51% | 88.26% | 70.35% | 62.05% | 66.22% | 98.12% | 94.32% | 97.68% |
| | | ✓ | 92.10% | 89.95% | **91.13%** | 78.49% | 77.14% | **74.82%** | 98.39% | 95.81% | **98.18%** |
| | Cascade R-CNN | ✗ | 88.97% | 94.16% | 88.67% | 69.19% | 83.80% | 68.00% | 98.12% | 98.12% | 98.00% |
| | | ✓ | 92.10% | 92.95% | **91.53%** | 78.49% | 82.32% | **76.32%** | 98.39% | 97.60% | **98.25%** |
| One-Satge | SSD | ✗ | 69.49% | 96.43% | 68.93% | 47.67% | 87.23% | 46.52% | 79.57% | 99.33% | 79.38% |
| | | ✓ | 78.12% | 93.82% | **76.59%** | 67.44% | 85.29% | **64.67%** | 83.06% | 97.48% | **82.44%** |
| | RetinaNet | ✗ | 72.43% | 94.26% | 71.51% | 48.26% | 83.00% | 45.66% | 83.60% | 97.80% | 83.35% |
| | | ✓ | 81.80% | 84.28% | **77.85%** | 68.02% | 61.26% | **58.06%** | 88.17% | 97.33% | **87.50%** |

## B. Training Details

In MMDetection (https://github.com/open-mmlab/mmdetection), BSTM is applied to four deep learning detectors (Faster R-CNN, Cascade R-CNN, RetinaNet and SSD). ResNet-50-FPN [24] is set as backbones of the first three and VGG-16 as backbone of SSD. All samples are resized into 512×512 [20] for network training. We train these four detectors for 12 epochs via Stochastic Gradient Descent (SGD) [28] with learning rate = 0.02, momentum = 0.9, weight decay = 0.001 and 4 batch size.

## C. Evaluation Indices

Recall, Precision and mAP are defined by [17]:

$$\text{Recall} = \text{TP} / (\text{TP} + \text{FN}), \quad \text{Precision} = \text{TP} / (\text{TP} + \text{FP}) \quad (4)$$

$$\text{mAP} = \int_0^1 P(R) \, \mathrm{d} R \quad (5)$$

where TP is True Positive, FN is False Negative, FP is False Positive and the full name of mAP is mean Average Precision.

## IV. RESULTS

### A. SAR Ship Detection Results

Fig. 6 shows the detection results of different methods on the test set of SSDD. Intersection over Union (IOU) threshold [17] is set as 0.5. From Fig. 6, when BSLM is adopted, the detection performance is universally improved, e.g., some false alarms are suppressed (marked in red ellipse), some missed detections are detected again (marked in red rectangle) and the confidence scores also universally become higher (marked in red arrow).

### B. Quantitative Performance Comparison

Table II shows the evaluation indices. Fig. 7 shows their P(R) curves. From Table II, the following conclusions can be drawn:
1) On the offshore and inshore scenes, BSLM improved Faster R-CNN, Cascade R-CNN, SSD and RetinaNet by 2.87%, 2.86%, 7.66% and 6.34% mAP, respectively.

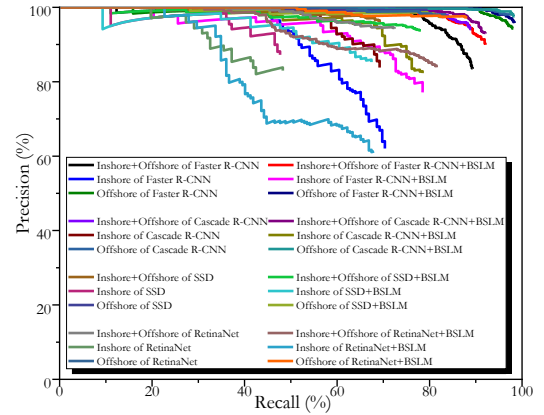

Fig. 7. P(R) curves of different methods. P is Precision and R is Recall.

2) On the inshore scenes, BSLM improved Faster R-CNN, Cascade R-CNN, SSD and RetinaNet by 8.60%, 8.32%, 18.15% and 12.40% mAP, respectively. Therefore, the learning representation ability of inshore scenes is greatly enhanced.
3) On the offshore scenes, BSLM improved Faster R-CNN, Cascade R-CNN, SSD and RetinaNet by 0.50%, 0.25%, 3.06% and 4.15% mAP, respectively. Notably, while enhancing inshore learning ability, offshore one does not become weak but a little stronger. One possible reason may be that without BSLM, networks may happen overfitting for too many offshore samples' over-learning, causing modest generalization ability, but with BSLM, networks can distract partial attention to the inshore, avoiding the above over-learning.
4) The accuracy gap between inshore and offshore still exists, but their gap is far lower than the original one.
5) When BSLM is applied to one-stage detectors, the accuracy gain is bigger than two-stage, so one-stage detectors' biggest poor accuracy defect is smoothly solved. To be clear, BSLM is merely used in training, not reducing detection speed.

TABLE III
EVALUATION INDICES OF DIFFERENT AUGMENTATION METHODS IN BSLM.

| Type | Method | Replicate | Noise | Rotate | Inshore + Offshore | | | Inshore | | | Offshore | | |
|------|--------|-----------|-------|--------|--------|-----------|-----|--------|-----------|-----|--------|-----------|-----|
| | | | | | Recall | Precision | mAP | Recall | Precision | mAP | Recall | Precision | mAP |
| Two-Stage | Faster R-CNN | ✓ | ✗ | ✗ | 92.10 | 89.95 | 91.13 | 78.49 | 77.14 | 74.82 | 98.39 | 95.81 | 98.18 |
| | | ✗ | ✓ | ✗ | 92.28 | 88.38 | 91.59 | 79.07 | 73.12 | 76.15 | 98.39 | 95.81 | 98.16 |
| | | ✗ | ✗ | ✓ | 92.10 | 86.68 | 91.30 | 78.49 | 66.50 | 74.91 | 98.39 | 97.60 | 98.21 |
| | | ✗ | ✓ | ✓ | 93.20 | 84.08 | **92.22%** | 81.98 | 62.67 | **76.59%** | 98.39 | 96.83 | **98.31%** |
| | Cascade R-CNN | ✓ | ✗ | ✗ | 92.10 | 92.95 | 91.53 | 78.49 | 82.32 | 76.32 | 98.39 | 97.60 | 98.25 |
| | | ✗ | ✓ | ✗ | 92.10 | 93.47 | 91.59 | 77.91 | 84.28 | 75.51 | 98.66 | 97.35 | **98.60%** |
| | | ✗ | ✗ | ✓ | 92.28 | 92.96 | **91.84%** | 78.49 | 81.82 | 76.24 | 98.66 | 97.87 | **98.60%** |
| | | ✗ | ✓ | ✓ | 92.28 | 92.79 | 91.73 | 79.07 | 80.47 | **76.33%** | 98.39 | 98.39 | 98.33 |
| One-Stage | SSD | ✓ | ✗ | ✗ | 78.12 | 93.82 | **76.59%** | 67.44 | 85.29 | **64.67%** | 83.06 | 97.48 | **82.44%** |
| | | ✗ | ✓ | ✗ | 71.14 | 94.16 | 70.05 | 60.47 | 86.67 | 58.36 | 76.08 | 97.25 | 75.46 |
| | | ✗ | ✗ | ✓ | 72.43 | 94.03 | 71.24 | 59.88 | 82.40 | 56.84 | 78.23 | 98.98 | 77.92 |
| | | ✗ | ✓ | ✓ | 71.51 | 94.65 | 70.69 | 58.14 | 84.75 | 55.94 | 77.69 | 98.63 | 77.43 |
| | RetinaNet | ✓ | ✗ | ✗ | 81.80 | 84.28 | **77.85%** | 68.02 | 61.26 | **58.06%** | 88.17 | 97.33 | **87.50%** |
| | | ✗ | ✓ | ✗ | 78.68 | 86.46 | 75.56 | 62.21 | 67.30 | 54.27 | 86.29 | 95.54 | 85.30 |
| | | ✗ | ✗ | ✓ | 79.41 | 82.13 | 74.79 | 63.37 | 57.37 | 51.81 | 86.83 | 96.13 | 86.08 |
| | | ✗ | ✓ | ✓ | 80.70 | 82.06 | 76.32 | 64.53 | 57.51 | 54.14 | 88.17 | 95.91 | 87.25 |

## C. Discussion on Different Augmentation Methods

Table III shows different augmentation methods' influences on model generalization ability where "replicate" is sample replication, "noise" is random gaussian noise addition with mean 0 and variance 0.1, and "rotate" is 90°, 180° and 270° rotation. From Table III, the following conclusions can be drawn:

1) For two-stage detectors, noise addition or rotation operation obtains better generalization ability than sample replication, in line with traditional machine learning common sense.

2) For one-stage detectors, these two methods do not obtain better accuracy than sample replication. One possible reason is that one-stage detectors may be sensitive to noise and rotation. Still, their overall detection performances both remain superior to initial detection performances of not using BSLM.

## V. CONCLUSIONS

BSLM is proposed for offshore and inshore SAR ship detection. First, GAN is used to extract image scene features. Then, K-means is used to perform a scene binary cluster. Finally, inshore samples are augmented to balance offshore ones. As a result, different scene learning bias is eased. Experimental results on SSDD verified the effectiveness of BSLM. BSLM is a universal mechanism that is also effective for more other detectors.

Our work of this letter is of great meaning, because:

1) We innovatively solved scene detection imbalance from the dataset perspective, even though simple but rather effective.

2) We originally used GAN for unsupervised SAR image scene feature extraction that can also stimulate new research ideas, e.g., scene adaption, scene recognition before detection, etc.